\documentclass[11pt, a4paper, logo, onecolumn]{google}

\usepackage[authoryear, sort&compress, round]{natbib}

\uselogo{} 

\title{Covariance-aware sampling for Diffusion Models.}

\correspondingauthor{arischioppa@google.com}

\usepackage{caption}
\usepackage{subcaption}

\usepackage{refcount}

\newcommand\soutpars[1]{\let\helpcmd\sout\parhelp#1\par\relax\relax}
\long\def\parhelp#1\par#2\relax{%
  \helpcmd{#1}\ifx\relax#2\else\par\parhelp#2\relax\fi%
}

\newcounter{parcounter}[section]
\renewcommand{\theparcounter}{\arabic{section}.\arabic{parcounter}}
\newcommand\numparagraph[1]{%
    \refstepcounter{parcounter}%
    \paragraph{\S\theparcounter\ #1}
}

\usepackage{color}

\definecolor{deepblue}{rgb}{0,0,0.5}
\definecolor{deepred}{rgb}{0.8,0,0}
\definecolor{deepgreen}{rgb}{0,0.5,0}
\definecolor{grey1}{rgb}{0.5,0.5,0.5}
\usepackage{listings}
\DeclareFixedFont{\ttb}{T1}{txtt}{bx}{n}{10} 
\DeclareFixedFont{\ttm}{T1}{txtt}{m}{n}{10}  

\newcommand\pythonstyle{\lstset{
		language=Python,
		basicstyle=\small\ttm,
		otherkeywords={self},             
		keywordstyle=\small\ttb\color{deepblue},
		emph={},          
		emphstyle=\small\color{deepred},    
		stringstyle=\small\color{deepgreen},
		commentstyle=\small\color{grey1}\ttm,
		frame=tb,                         
		showstringspaces=false,            %
		breaklines=true,
		tabsize=2,
        numbers=left,
        numberstyle=\tiny,
        breaklines=true,
        escapeinside={(*@}{@*)}
}}
\lstnewenvironment{python}[1][] {
	\pythonstyle
	\lstset{#1}
}{}

\newcommand\pythoninline[1]{{\pythonstyle\lstinline!#1!}}

\usepackage{graphicx}
\usepackage{pgffor}
\usepackage{pgfkeys}
\usepackage{xspace} 

\def\convDCT{ConvDCT\xspace}
\def\condMean#1;#2.{{\operatorname{E}}[#1 | #2]}
\def\condCov#1;#2.{{\operatorname{Cov}}[#1 | #2]}

\author[1]{Andrea Schioppa}
\author[2]{Tim Salimans}

\affil[1]{GDM - Amsterdam}
\affil[2]{work done at GDM - Amsterdam}

\begin{abstract}
We present a covariance-aware sampler that improves the quality of pixel-space Diffusion Model (DM) sampling in the few-step regime.
We hypothesize that in the few-step regime samplers fail because they rely solely on the predicted mean of the reverse distribution, while our solution explicitly models the reverse-process covariance. Our method combines Tweedie's formula to estimate the covariance with an efficient, structured Fourier-space decomposition of the covariance matrix. Implemented as an extension of DDIM, our method requires only a minimal overhead: one extra Jacobian-Vector Product (JVP) per step.
We demonstrate that for pixel-based DMs, our method consistently produces superior samples compared to state-of-the-art second order samplers (Heun, DPM-Solver++) and the recent aDDIM sampler, at an identical number of function evaluations (NFE).
\end{abstract}

\begin{document}

\maketitle
\addtocontents{toc}{\protect\setcounter{tocdepth}{0}}
\section{Introduction}

Diffusion Models (DMs) have achieved state-of-the-art results on diverse tasks, including image generation~\citep{saharia2022photorealistic, rombach2022high, ho2022cascaded, ho2020denoising}, video generation~\citep{openai2024sora, blattmann2023stable, ho2022imagen}, audio~\citep{liu2023audioldm, zhu2023edmsound, kong2020diffwave, chen2020wavegrad}, data compression~\citep{yang2025universal, yang2023lossy, theis2022lossy}, and inverse problems~\citep{Rissanen2025FreeHunch, boys2024tweediemomentprojecteddiffusions, chung2023diffusion}. A key characteristic of DMs is their gradual sampling procedure, which transforms simple Gaussian noise into a complex data sample.  However, generating high-quality samples requires a large number of steps, making inference slow. Extensive research has focused on reducing this computational burden, generally falling into two categories. The first is \emph{distillation}, which involves training a new, faster model from a pre-trained DM~\citep{salimans2024multistep, song2023consistency, salimans2022progressive}. The second is the \emph{development of higher-order samplers}, which improve numerical integration by viewing the sampling process as solving an Ordinary or Stochastic Differential Equation (ODE/SDE)~\citep{Karras2022ElucidatingDesignSpace, DPM-Solver++, GenieSampler}.

In this paper, we propose a new sampler inspired by recent work on inverse problems and higher-order score matching. While inspired by higher-order methods, our sampler is a first-order, non-deterministic method that adds noise at each step. To motivate our approach, we observe that while the true reverse conditional distribution $p(x_{t-1} | x_t)$ is intractable, its moments can be estimated via Tweedie's formula. Prior work has used second-order moment information (i.e., the covariance) to improve the \emph{training objective}~\citep{Meng2021HighOrder, Lu2022MaximumLikelihood}. We instead propose to leverage an estimate of the conditional covariance, $C_t = \condCov x_{t-1};x_t.$, directly during the \emph{sampling process} without modifying the training objective. This yields a first-order sampler that adds structured noise based on $C_t$ at each step. Directly estimating the large covariance matrix $C_t$ at each step would be computationally prohibitive. To make our sampler competitive, we develop a method to approximate $C_t$ using only a single additional function evaluation per step. This is achieved by combining two ingredients: Hutchinson's trace estimator and a structured decomposition of $C_t$ in the frequency domain. 
This decomposition involves two key design aspects: the block-wise structure assumed for the covariance and the choice of mapping to the frequency domain. Regarding the latter, we exploit the fact that pixel-space frequency decompositions maintain a clear structural and interpretable correspondence to spatial features, a property that might degrade within the abstract representations of compressed latent spaces.
Our experiments demonstrate that the optimal block-wise structure requires granular modeling at each spatial location. For the frequency mapping, we find the Discrete Cosine Transform (DCT) yields the best results. For regimes with very few steps, we introduce a convolutional version of the DCT—a novel contribution of this work—which performs even better.

Empirically, we find that for pixel-space diffusion models, our sampler yields better results than state-of-the-art second-order samplers (Heun~\citep{Karras2022ElucidatingDesignSpace}, DPM-Solver++~\citep{DPM-Solver++}) and the recent aDDIM~\citep{heek2024multistepconsistencymodels} sampler for the same number of function evaluations (NFE).

\section{Method}
\label{sec:method}

\numparagraph{Intuition}
We hypothesize that accurately estimating the variance of the reverse-process posterior $P(x_{t-1} | x_t)$ is crucial for high-quality sampling when using fewer steps than the training configuration. With many steps, the transitions between $x_t$ and $x_{t-1}$ are small, and the conditional mean $\condMean x_{t-1};x_t.$ is a sufficient statistic for generating good samples. However, as the number of steps decreases, these transitions become larger, and relying solely on the mean leads to a collapse in sample diversity. The core challenge is that while the forward process $P(x_t | x_{t-1})$ is a tractable Gaussian, the posterior $P(x_{t-1} | x_t)$ is not. Existing methods circumvent this by adding noise based on heuristics or data-driven estimates \citep[Sec.~3.2]{heek2024multistepconsistencymodels}; in particular, the aDDIM sampler
introduced in \citep{heek2024multistepconsistencymodels} estimates the conditional variance of $x_{t-1}$ given $x_t$ using  a batch of training data and then uses it to add noise to the
estimate prediction of $x_{t-1}$ given by DDIM; aDDIM is shown to out-perform DDIM in pixel-space models.

In contrast, we propose a more principled approach. Although the full posterior is intractable, its covariance can be computed analytically via Tweedie's formula~\citep{efron_tweedie}. To manage the high dimensionality of the covariance matrix, we approximate it as a diagonal matrix in the Fourier domain. This allows us to sample a unique variance for each Fourier component, a choice motivated by prior work demonstrating that the success of diffusion models on images is deeply connected to their ability to model structure in the frequency domain~\citep[Sec.~2.2]{Rissanen2023InverseHeat}, \citep{dieleman2024spectral}.
Moreover, adjusting the guidance levels for different Fourier frequencies has been shown to improve generation quality~\citep{sadat2025guidance}.

\numparagraph{Mathematical derivation}
Since the forward transition $P(x_t | x_{t-1})$ is Gaussian, the reverse posterior $P(x_{t-1} | x_t)$ is a member of the exponential family. For distributions of this type, Tweedie's formula~\citep{efron_tweedie} connects the distribution moments to derivatives of the log-partition function. Concretely, we can use Bayes' rule to express $P(x_{t-1} | x_t)$ as:

\begin{equation}\label{eq:bayes}
    P(x_{t-1} | x_t) = \exp(\lambda x_{t-1}^T x_t - F(x_t) - G(x_{t-1})) P(x_{t-1}),
\end{equation}
where $F$, $G$ are functions and $\lambda$ is a constant. More details regarding~\eqref{eq:bayes} can be found in Appendix~\ref{appx:math-details-bayes}.
Taking the gradient and the Hessian of~\eqref{eq:bayes} wrt.~$x_t$ and observing that the expectation over $x_{t-1}$
of the gradient and the Hessian is zero as $\int P(x_{t-1} | x_t)\,dx_{t-1} = 1$ is constant in $x_t$, one gets the following formulas for the conditional mean and covariance:

\begin{align}
    \lambda \condMean x_{t-1};x_t. &= \nabla_{x_t}F(x_t), \label{eq:tweedie_mu}\\
    \lambda^2 \condCov x_{t-1};x_t. &= \nabla^2_{x_t}F(x_t). \label{eq:tweedie_sig}
\end{align}

We define a function $M(x_t, \theta)$ with neural network parameters $\theta$ to predict the conditional mean $\condMean x_{t-1};x_t.$; this mean is a deterministic function of the neural network's output, the form of which depends on the chosen parameterization of the diffusion process. Using~\eqref{eq:tweedie_mu} and~\eqref{eq:tweedie_sig} we arrive at

\begin{equation}\label{eq:model_covariance}
    \condCov x_{t-1};x_t. = \nabla_{x_t} M(x_t, \theta).
\end{equation}

While computing the full Jacobian matrix $\nabla_{x_t} M(x_t, \theta)$ is often prohibitively expensive, this formulation is computationally practical because it allows for the efficient calculation of Jacobian-vector products (JVPs). Computing a JVP such as $(\nabla_{x_t} M(x_t, \theta))\cdot \varepsilon$ for a vector $\varepsilon$ only requires a single additional forward pass through the network.

\numparagraph{Structured Covariance in the Fourier Domain}
The full covariance matrix, $C_t = \condCov x_{t-1};x_t.$, has dimensions $D\times D$, making it computationally intractable to compute or sample from for high-dimensional data $x_t$. A common simplification is to approximate $C_t$ with an isotropic matrix $t_t I_D$, where the scalar variance $t_t$ can be computed using a trace estimator. We propose a more expressive, structured approximation. Instead of assuming independence in the $x_t$-domain, we model it in a frequency domain by approximating the covariance as a block-diagonal matrix $\hat{C}_t$ in a suitable basis:
\begin{equation}\label{eq:diag_blocks}
    \hat{C}_t = \operatorname{diag}(c_{1,t}I_{D_1}, \dots, c_{k,t}I_{D_k}).
\end{equation}
Here, each scalar variance $c_{i,t}$ is shared across a group of $D_i$ frequency components. Each $c_{i,t}$ can be estimated efficiently using Hutchinson's trace estimator~\citep{Hutchinson_original, hutchinson_modern_analysis}, which leverages the JVPs of $\nabla_{x_t} M(x_t, \theta)$ discussed previously.

The choice of frequency transform is critical for this approximation. We evaluated several options, including the Haar~\citep{Haar1910} and DWT (5/3) wavelets~\citep{LeGall1988Subband}, the DCT~\citep{DCT_original} applied to $8\times 8$ blocks, and a convolutional version of the DCT we term \convDCT. Our empirical results show that \convDCT and DCT consistently yields superior performance, with the former being preferable in  setups with fewer function evaluations. This \convDCT operator works by sliding a DCT kernel over the image's $8\times 8$ blocks. The variances for the resulting 64 frequency components are then averaged across all channels.

\numparagraph{Implementation}
A Python implementation for the \convDCT and our sampler is available in Appendix~\ref{appx:impl-details}. A crucial aspect of our implementation is to apply the JVP only to the guided $x$-prediction; this
saves a fourth function evaluation.
\section{Related Work}
\label{sec:related_work}

\numparagraph{Samplers}
The computational cost of sampling in Diffusion Models remains a crucial area of research, primarily focusing on reducing the required number of integration steps. This effort has led to the development of sophisticated higher-order samplers, notably Heun’s method~\citep{Karras2022ElucidatingDesignSpace} and DPM-Solver++~\citep{DPM-Solver++}.
Other work has explored alternative numerical approaches. For instance, \citep{GenieSampler} introduced a second-order sampler based on the truncated Taylor method. Their technique is structurally similar to the second-order DPM-Solver++~\citep{DPM-Solver++}, but it estimates the necessary second-order correction by leveraging an extra JVP (Jacobian-Vector Product) step to compute the time derivative of the error prediction. In a different vein, \citep{heek2024multistepconsistencymodels} introduced the aDDIM sampler, motivated by multi-step consistency models. This is a first-order sampler that uses a heuristic to estimate the posterior covariance $p(x_{t-1} | x_t)$. In contrast, our method is also a first-order sampler, but it estimates this crucial covariance term using a direct formulation based on Tweedie's formula.

\numparagraph{Higher Order score matching}
A separate line of inquiry focuses on enhancing the score model training objective itself. \citep{Meng2021HighOrder} proposed learning a second-order score model alongside the standard first-order model, with the second-order loss being explicitly derived via Tweedie's formula. While the resulting network produces superior samples, this approach introduces significant computational challenges, particularly when scaling to higher image resolutions. Similarly, \citep{Lu2022MaximumLikelihood} demonstrated a theoretical discrepancy between the objectives used to maximize the likelihood of score-based diffusion models when sampling via SDEs versus ODEs. While the SDE likelihood can be upper-bounded by the standard score matching objective, the ODE likelihood requires optimizing a more complex objective that involves higher-order moments of the distribution.
More recently, \citep{ou2025improving} propose learning a second-order moment network after the primary score model has been trained. This is achieved by feeding the activations of the first-order model into a smaller auxiliary network to produce a diagonal approximation of the covariance.
Ultimately, these approaches \citep{Meng2021HighOrder, Lu2022MaximumLikelihood, ou2025improving} require either fundamental modifications to the training pipeline or additional post-hoc training stages, which can limit their scalability and ease of use. In contrast, our work focuses exclusively on the sampling procedure. We offer a plug-in enhancement that requires no changes to the model training or architecture and leverage a structured decomposition of the covariance matrix in Fourier space.

\numparagraph{Inverse Problems}
Diffusion Models have rapidly become a powerful tool for solving Inverse Problems, such as reconstructing an image $x_0$ from noisy observations $y$. A central challenge in this domain is accurately approximating the conditional distribution $p(y|x_t)$.To address this, a significant body of recent work has leveraged Tweedie's formula to better estimate the covariance matrix $\Sigma_t$ and thus improve the approximation of $p(y|x_t)$. These approaches differ primarily in their structural assumptions about $\Sigma_t$. For instance, \citep{peng2024improving} introduced the assumption that $\Sigma_t$ is diagonal in the DWT (5/3) domain, while \citep{Rout2024Tweedie} adopted a simpler assumption that $\Sigma_t$ is merely a multiple of the identity matrix. Moving beyond diagonal or scalar approximations, \citep{boys2024tweediemomentprojecteddiffusions} utilized a row-sum approximation of $\Sigma_t$. For linear inverse problems, the Free Hunch method~\citep{Rissanen2025FreeHunch} further refines this by combining a low-rank covariance estimate with low-rank BGFS updates for optimization, while~\citep{yismaw2025covariance} proposes a covariance-correction term that can be computed without back-propagating through the model. \citep{zheng2025wavelet} shows that performing the posterior update in wavelet (DWT) space leads to better samples. Regarding covariance approximations, our work extends these efforts by systematically investigating various techniques, including different Fourier mappings into frequency domains and detailed block-wise decompositions of the $\Sigma_t$ matrix.

\section{Experiments}
\label{sec:experiments}

\begin{figure}
    \centering
    \includegraphics[width=0.8\linewidth]{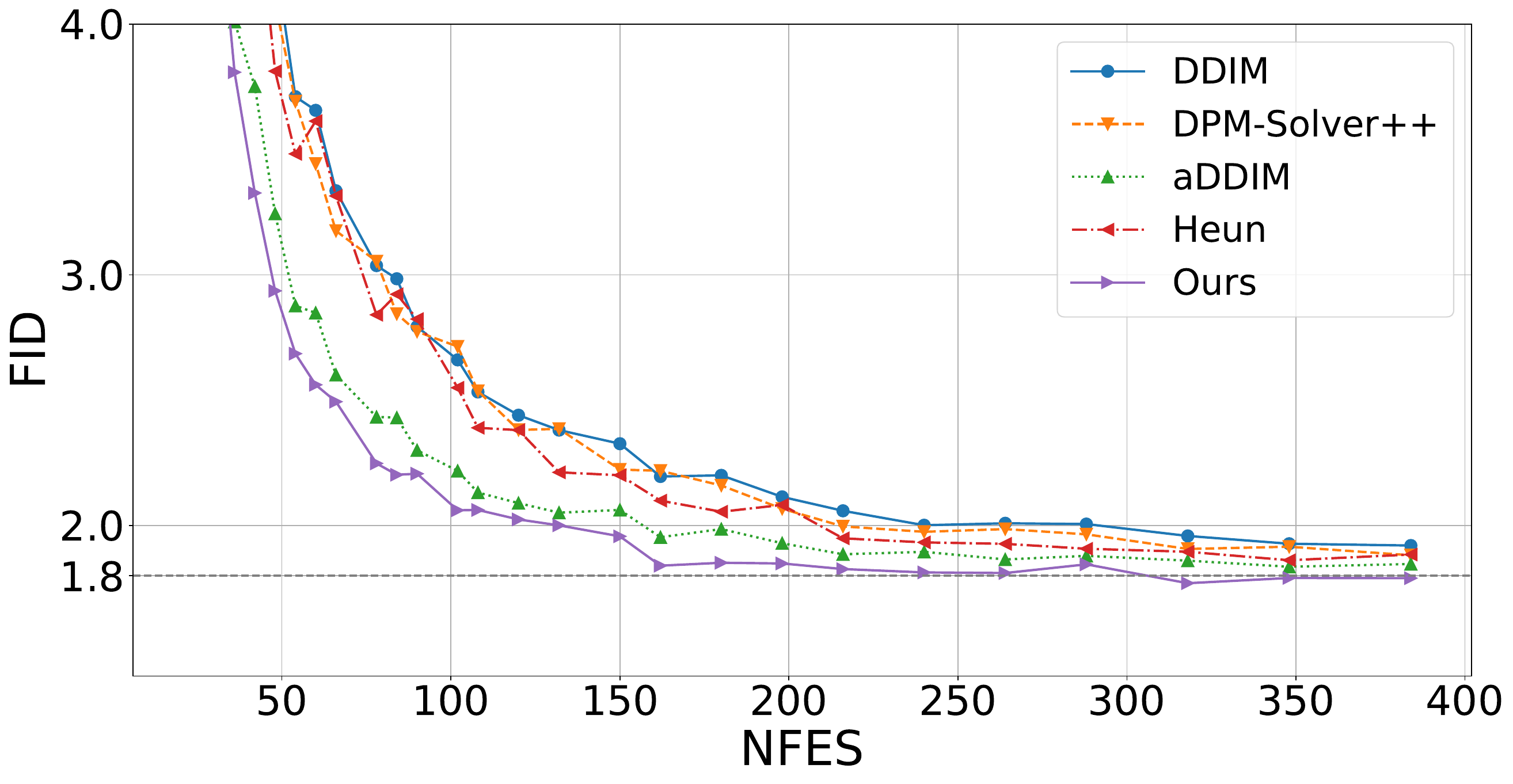}
    \caption{Our method outperforms the second-order samplers Heun and DPM-Solver++ and the aDDIM for the same number of function evaluations (NFEs). For a sufficient number of steps, all methods converge to the same FID($@50k)$. Note that in this setup the DDIM sampler is the less performant.}
    \label{fig:sid2}
\end{figure}

\numparagraph{Pixel-space Model}
We train a diffusion model in pixel space using the setup described in SiD2~\citep{SimplerDiffusion} on Imagenet512. The model achieves a competitive FID (we report FID at 50k samples) of 1.6 using 512 sampling steps with guidance and the DDPM sampler. Specifically, this model is trained with a sigmoid loss and a shifted cosine schedule as detailed in~\citep[Appendix.~B]{SimplerDiffusion}.
In the same experimental setup, but using the DDIM sampler, the model achieves an FID of 1.8.  Following~\citep{SimplerDiffusion} we found that a guidance of 1.2 and a guidance interval on logsnr $(-3, +5)$~\citep{guidance2024interval} yielded the best results; we keep these guidance parameters fixed in our experiments.
The number of function evaluations (NFEs) per sampling step varies based on the sampler and guidance method. For first-order samplers like DDIM or aDDIM~\citep{heek2024multistepconsistencymodels}, each step corresponds to 2 function evaluations due to guidance. In contrast, second-order samplers such as Heun~\citep[Alg.~1]{Karras2022ElucidatingDesignSpace} and DPM-Solver++ \citep{DPM-Solver++} require 4 function evaluations per step. Our proposed sampler, however, requires only 3 function evaluations per step, as the JVP is applied exclusively to the guided noise. As shown in Figure~\ref{fig:sid2}, our sampler consistently outperforms the others across various NFEs, although all methods converge to a similar FID as the number of steps becomes sufficiently large.

\begin{figure}
\begin{subfigure}[b]{0.5\linewidth}
    \centering
    \includegraphics[width=\linewidth]{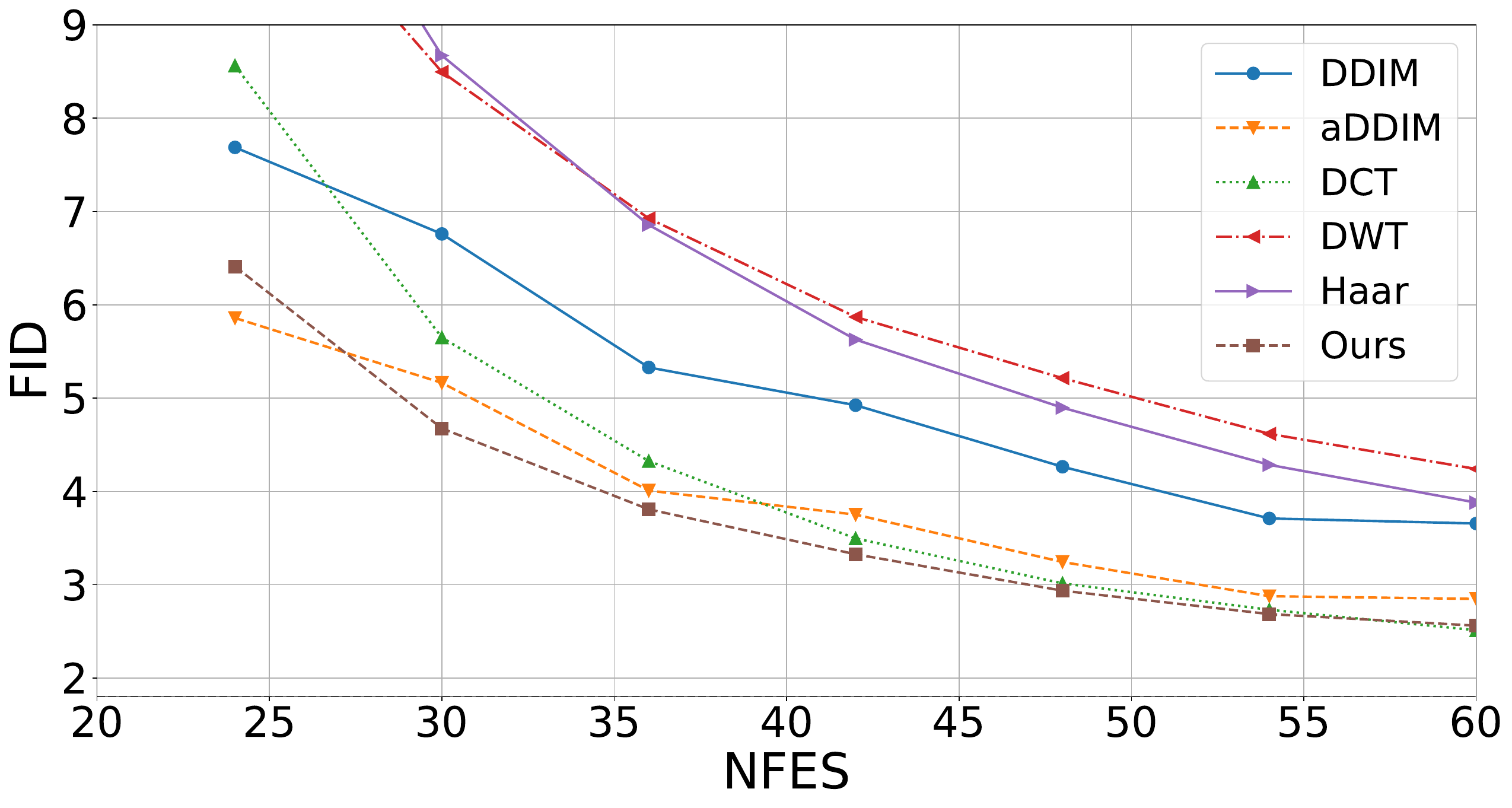}
\end{subfigure}
\begin{subfigure}[b]{0.5\linewidth}
    \centering
    \includegraphics[width=\linewidth]{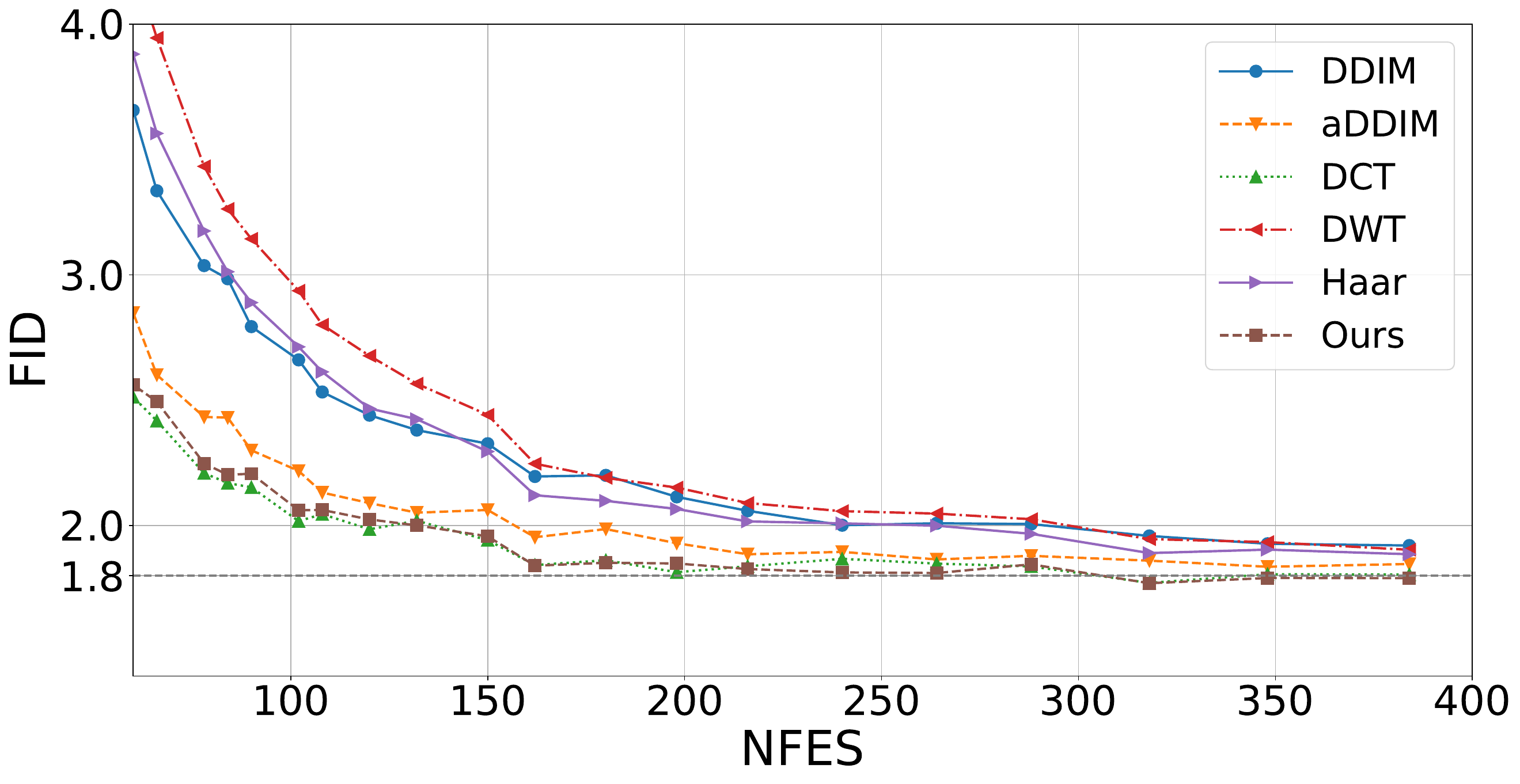}
\end{subfigure}
\caption{Both \convDCT and DCT consistently outperforms other mappings to Fourier space. For few NFEs \convDCT is better (left) while for more NFEs they are on par (right).}
\label{fig:fourier_ablation}
\end{figure}

\begin{figure}
    \centering
    \includegraphics[width=0.8\linewidth]{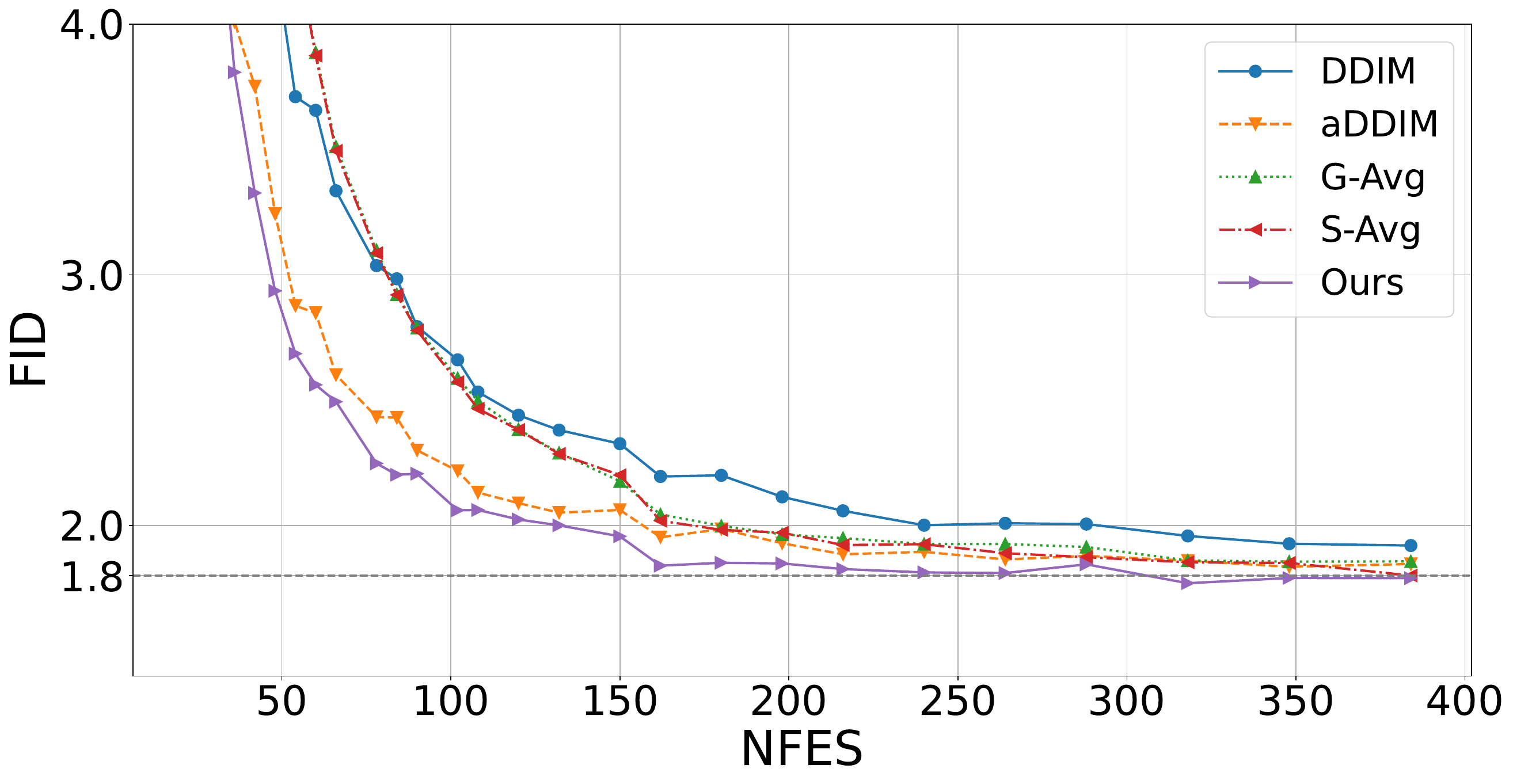}
    \caption{The choice of average has a significant effect on performance.}
    \label{fig:average_ablation}
\end{figure}

\numparagraph{Ablations on the Fourier Transform and the averaging.}
In this section, we present an ablation study on the choice of the Fourier transform and the averaging for our method. We evaluate the performance of our proposed convolutional version, \convDCT, against several standard transforms: the Haar and DWT (5/3) wavelets, and the block-based $8\times 8$ Discrete Cosine Transform (DCT). Our \convDCT method considers the neighborhood of each pixel, which is not the case for the other transforms. For a fair comparison, the number of levels for the Haar and DWT wavelets were chosen to match the number of Fourier components produced by the $8\times 8$ DCT. As illustrated in Figure~\ref{fig:fourier_ablation}, \convDCT and DCT consistently outperforms the other transforms, although the performance difference diminishes as the number of NFEs increases (and thus, the step size decreases). \convDCT is more
advantageous than DCT for a small number of function evaluations, while DCT and \convDCT are on par with a higher number of function evaluations. Regarding the 
averaging, this corresponds on the assumptions on the structure of $\hat{C}_t$ make in equation~\ref{eq:diag_blocks}; with a global average (G-Avg) $\hat{C}_t$
would be assumed to be decomposed into different blocks for each Fourier component, while with a spatial average (S-Avg) each $c_{i,t}$ each block would correspond
to a Fourier component and channel. We see in Figure~\ref{fig:average_ablation} that these averages have a worse performance than averaging just on the channels (Ours). Averaging just on the channels corresponds in assuming that for each pixel and spatial component we have a different $c_{i,t}$; this results in a more granular decomposition of $\hat{C}_t$
and appears crucial for a good performance.

\numparagraph{Application to Stable Diffusion}
In the setting of Stable Diffusion~\citep{rombach2022high} our sampler is not competitive with the deterministic DDIM sampler. We have included results in Appendix~\ref{sec:app_stable_diffusion}:
our investigation suggests that in this setting the Autoencoder is sensitive to noisy latents and this finding might be specific to the way that Autoencoder was trained.

\begin{figure}
    \centering
    \includegraphics[width=0.8\linewidth]{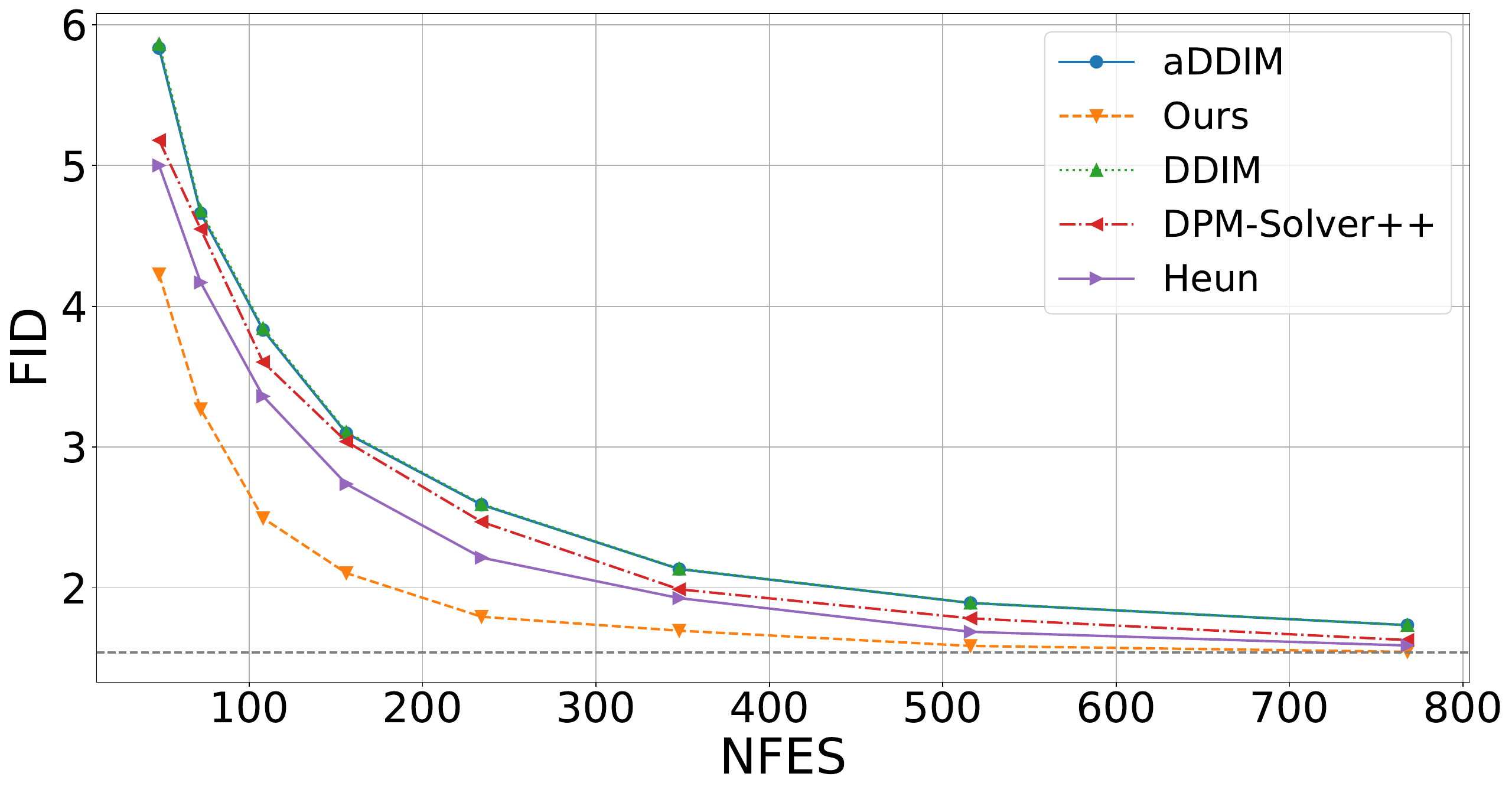}
    \caption{Sampling performance on ImageNet-128. Our method outperforms second-order samplers, while aDDIM and DDIM show similar performance, trailing behind second-order methods.}
    \label{fig:uvit_128}
\end{figure}

\numparagraph{Experiments on UViT and Imagenet128}
In this experiment, we evaluate our approach using a standard, pixel-space UViT configuration. We train a 400M parameter UViT model on ImageNet-128 for 500k iterations with a total batch size of 2048. The model utilizes a vanilla cosine schedule and $v$-parameterization. Using standard DDPM, we achieve an FID@50k of 1.48 with 512 sampling steps and a guidance scale of 1.2. As shown in Figure~\ref{fig:uvit_128}, aDDIM and DDIM exhibit comparable performance, both of which are surpassed by second-order samplers at equivalent Number of Function Evaluations (NFEs). Notably, our proposed method consistently outperforms these second-order baselines.

\begin{figure}
\begin{subfigure}[b]{0.5\linewidth}
    \centering
    \includegraphics[width=\linewidth]{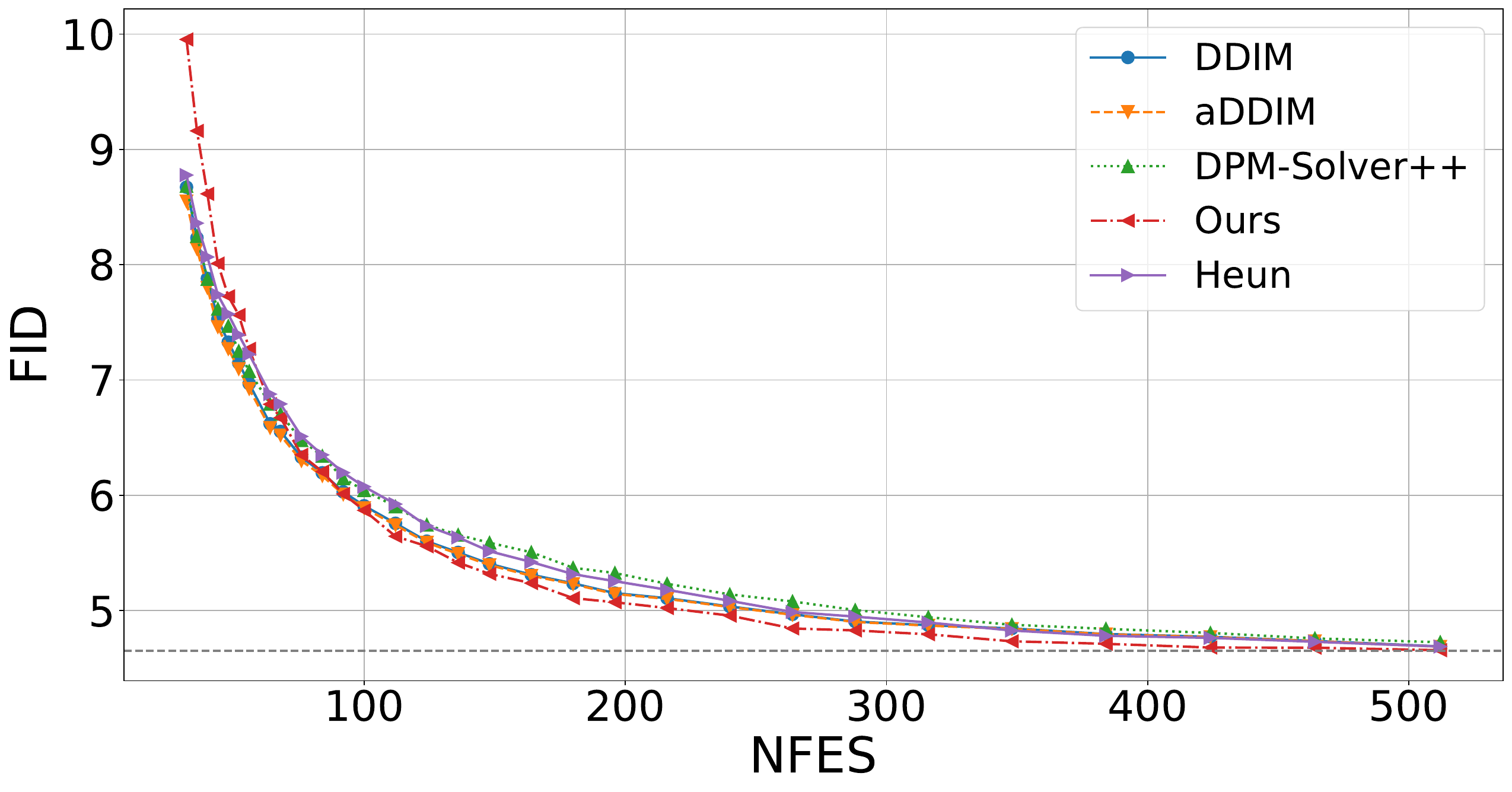}
    \subcaption{}
    \label{fig:cifar10_left}
\end{subfigure}
\begin{subfigure}[b]{0.5\linewidth}
    \centering
    \includegraphics[width=\linewidth]{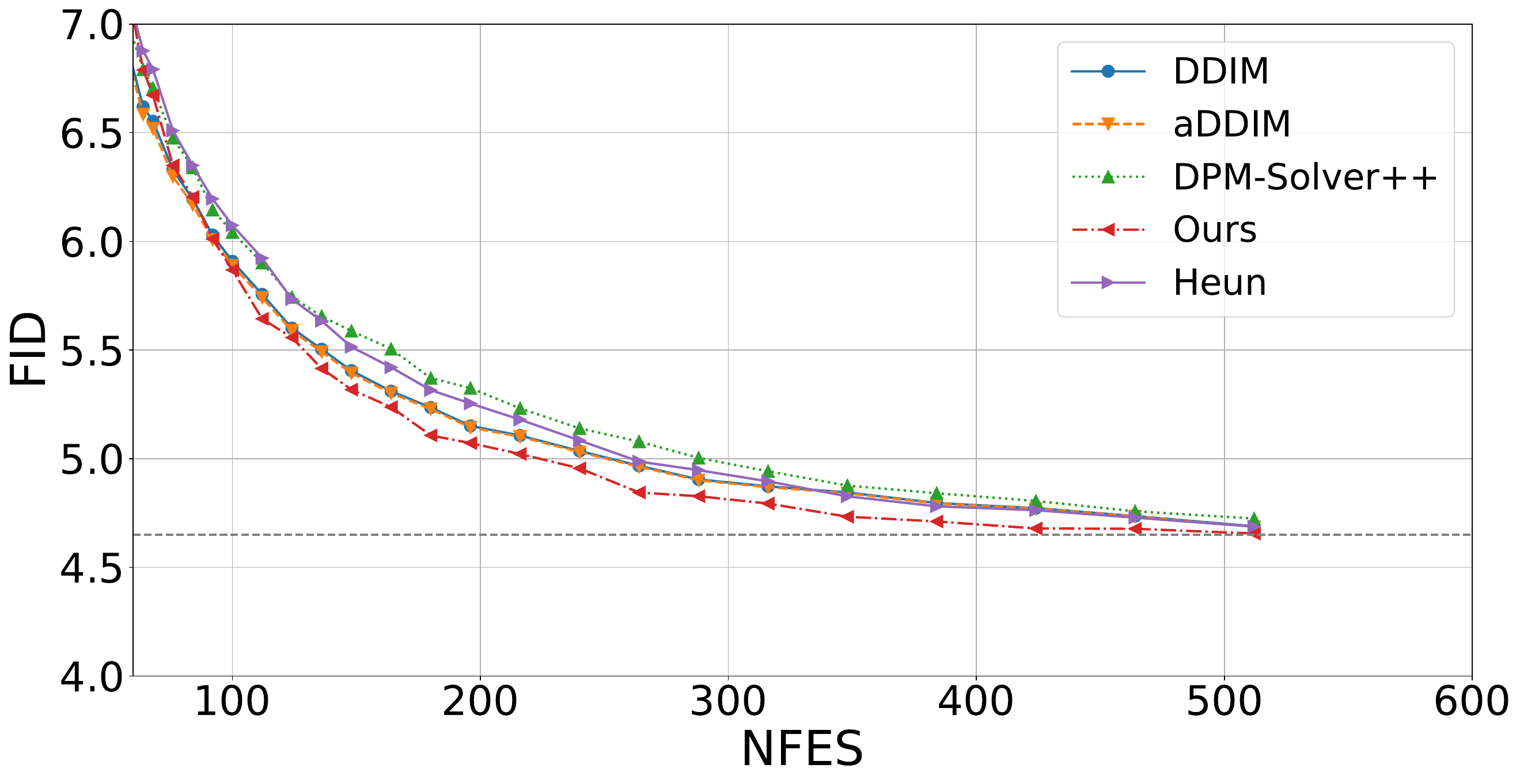}
    \subcaption{}
    \label{fig:cifar10_right}
\end{subfigure}
\caption{Comparison of samplers for unconditional CIFAR-10 generation. Left: performance across the full range of NFEs. Right: focused view of the high-NFE regime, where our method matches or exceeds aDDIM and DDIM performance despite the step-count penalty.}
\label{fig:cifar10}
\end{figure}

\numparagraph{Experiments for Unconditional Generation}
Finally, we evaluate unconditional generation on CIFAR-10. We train a 373M pixel-space UViT for 17k steps with a batch size of 2048, employing a cosine schedule and $v$-parameterization. This setup achieves a baseline DDPM FID@50k of 3.98 at $32\times 32$ resolution. In this comparison, second-order methods (Heun, DPM-Solver++) and our sampler require two function evaluations (NFEs) per sampling step, whereas aDDIM and DDIM require only one.  While our method is less performant in the extremely low-step regime (Figure~\ref{fig:cifar10_left}) -- a regime characterized by high absolute FID scores -- it significantly improves as the number of steps increases (Figure~\ref{fig:cifar10_right}), eventually outperforming all other samplers. Interestingly, aDDIM and DDIM outperform Heun and DPM-Solver++ in this specific setup. This suggests a challenging scenario for our method; despite being ``penalized" by having half the total steps of aDDIM/DDIM for a fixed NFE, our sampler still matches or exceeds their performance given a sufficient computational budget.

\section{Conclusions}
In this work, we introduced a novel first-order sampler for Diffusion Models that enhances samples quality by explicitly leveraging covariance information at each step. This covariance is derived directly from the trained model. To make the calculation of the large covariance matrix tractable, we combined two key techniques: a structured decomposition of the covariance in the frequency domain and a trace estimator employing Jacobian-Vector Products (JVP). Our approach consistently outperforms strong baselines. Specifically, our covariance-aware sampler achieves superior results compared to the aDDIM samplers and state-of-the-art second-order samplers (e.g., Heun, DPM-Solver++) for the same number of function evaluations (NFE).

While highly effective in the pixel space, we did not investigate thoroughly our sampler in the setting of Latent Diffusion Models (LDMs); while in the Appendix~\ref{sec:app_stable_diffusion} we have included a negative
result for the Autoencoder of Stable Diffusion~\citep{rombach2022high}, a more thorough investigation in the setting of LDMs is needed, which we leave for future work.

\section{Acknowledgements}
We thank Miloš Stanojević and Mirella Lapata for thoughtful feedback on our draft.

\bibliography{main}

\newpage
\appendix
\addtocontents{toc}{\protect\setcounter{tocdepth}{2}}
\tableofcontents

\section{Appendix: Application to Stable Diffusion}
\label{sec:app_stable_diffusion}
\begin{figure}
    \centering
    \includegraphics[width=0.8\linewidth]{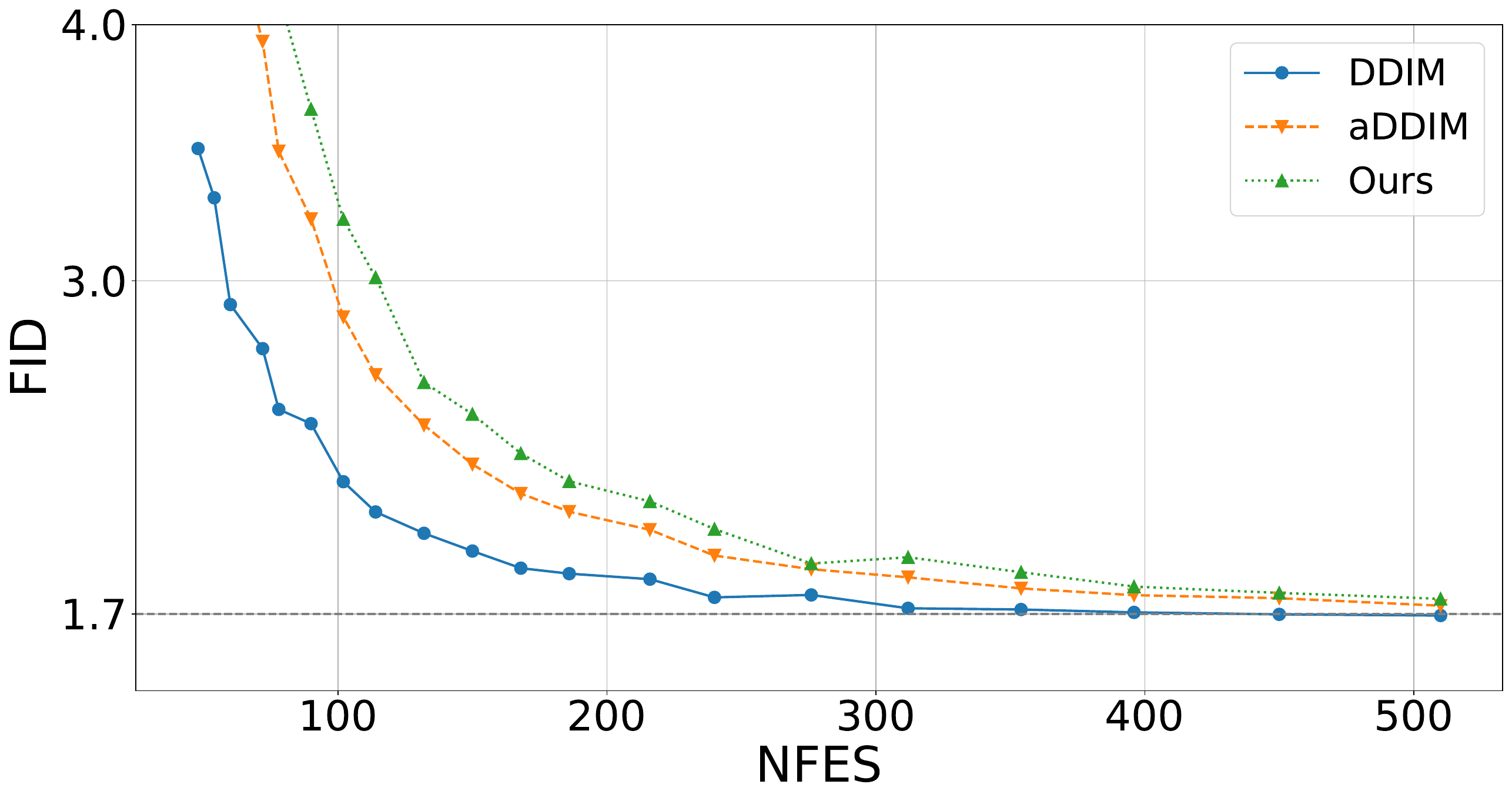}
    \caption{DDIM outperforms methods that add noise during sampling, such as the aDDIM and our covariance-aware sampler.}
    \label{fig:stable_diffusion}
\end{figure}

We apply our proposed method to a Latent Diffusion Model trained on Imagenet512, using an Autoencoder identical to the one in Stable Diffusion~\citep{Rombach2022LatentDiffusion}. Our model achieves an FID of 1.7 with 256 sampling steps. Interestingly, we observed that our method performed worse than the aDDIM sampler in this context. We also found that capping the noise introduced by our covariance-aware sampler improved the results, leading us to hypothesize that for this kind of auto-encoder adding noise to the latent space might in
general degrade the sampling performance. To verify this, we compared our method and the aDDIM against the DDIM sampler, which does not add noise. As shown in Figure~\ref{fig:stable_diffusion}, the DDIM sampler consistently outperforms the other methods: we conjecture that this is due to the decoder being highly sensitive to noisy latents.
However, this finding might be limited to the specific Autoencoder that we used, and we leave the extension of our sampler to Latent Diffusion Models for future work.

\section{Appendix: Further mathematical details}
\label{appx:math-details}
\subsection{Derivation of parameters \texorpdfstring{$\lambda$, $F$, $G$ in~\eqref{eq:bayes}}{lambda, F, G in the posterior equation}}
\label{appx:math-details-bayes}
We provide the explicit derivation of~\eqref{eq:bayes} to clarify the functional forms of $\lambda$, $F(x_t)$, and $G(x_{t-1})$.
First, recall the transition probability for the forward process:
\begin{equation}
    P(x_t | x_{t-1}) = C(t) \exp(-\frac{1}{2} \frac{\|x_t-\tilde\alpha(t, t-1) x_{t-1}\|^2}{\sigma^2(t)}),
\end{equation}
where $\sigma(t)$ is the noise level at time $t$. The coefficient $\tilde{\alpha}(t, t-1) = \alpha(t)/\alpha(t-1)$ 
ensures that $x_t$ maintains the correct mean relative to the signal schedule $\alpha(t)$.
Applying Bayes' rule, the posterior is given by:
\begin{equation}
    P(x_{t-1} | x_t) = P(x_t | x_{t-1}) \frac{P(x_{t-1})}{P(x_t)}.
\end{equation}
Taking the logarithm and expanding the quadratic term, we obtain:
\begin{equation}
\begin{split}
    \log\frac{P(x_{t-1} | x_t)}{P(x_{t-1})} &= \log C(t) -\frac{1}{2}\frac{\|x_t-\tilde\alpha(t) x_{t-1}\|^2}{\sigma^2(t)}+\log P(t) \\
    &= \log C(t) -\frac{1}{2}\frac{\|x_t\|^2}{\sigma^2(t)} -\frac{1}{2}\frac{\tilde\alpha^2(t, t-1)\|x_{t-1}\|^2}{\sigma^2(t)} \\
    &\quad+x_{t-1}^T x_t \frac{\tilde\alpha(t,t-1)}{\sigma^2(t)} + \log P(t) \\
    &=\underbrace{\frac{\tilde\alpha(t,t-1)}{\sigma^2(t)}}_{\lambda} x_{t-1}^T x_t  - \underbrace{\frac{1}{2}\frac{\tilde\alpha^2(t, t-1)\|x_{t-1}\|^2}{\sigma^2(t)}}_{G(x_{t-1})} \\
    &\quad - \underbrace{\left(-\log C(t) + \frac{1}{2}\frac{\|x_t\|^2}{\sigma^2(t)}  - \log P(t)\right)}_{F(x_t)}.
\end{split}
\end{equation}

\subsection{Adaptation to \texorpdfstring{$x$}{x}-prediction models}
\label{appx:math-details-x-pred}
While our primary experiments utilize $v$-prediction networks, our sampler implementation converts these to $x$-prediction for internal updates. This section justifies the identities used in Appendix~\ref{appx:impl-details}.
Assuming a variance-preserving (VP) schedule where $\alpha^2(t) + \sigma^2(t) = 1$, the clean data prediction $x_0$ relates to the posterior mean and covariance as follows:
\begin{align}
    \alpha(t-1) \condMean x_0;x_t. &= \condMean x_{t-1};x_t. \\
    \alpha^2(t-1) \condCov x_0;x_t. &= \condCov x_{t-1};x_t..
\end{align}
Following the Tweedie-style identities~\eqref{eq:tweedie_mu}, \eqref{eq:tweedie_sig} for Gaussian denoisers, these moments can be expressed in terms of $F(x_t)$:
\begin{align}
    \frac{\alpha(t)}{\sigma^2(t)} \condMean x_0;x_t. &= \nabla_{x_t}F(x_t) \\
    \frac{\alpha^2(t)}{\sigma^4(t)} \condCov x_0;x_t. &= \nabla^2_{x_t}F(x_t).
\end{align}

In the implementation (Listing~\ref{snippet:sampler}, line~\ref{line:noise_fac}), we utilize the relationship:
\begin{equation}
     \condCov x_0;x_t. = \frac{1}{\sigma(t)}\frac{\sigma(t)}{\alpha(t)} \nabla_{x_t}\condMean x_0;x_t..
\end{equation}
Specifically, we define the log signal-to-noise ratio (log-SNR) as $l$. Then $\sigma(t)/\alpha(t) = \exp(-l/2)$. Under the VP formulation, $\sigma^{-2}(t) = 1 + \exp(l)$
which allows for efficient computation using a softplus operation: $\sigma^{-1}(t) = \exp(-\text{softplus}(l)/2)$.

\section{Appendix: Implementation details}
\label{appx:impl-details}

Please refer to Appendix~\ref{appx:math-details-x-pred} as we work with the $x$-prediction for samplers.

We rely on the following libraries.
\begin{python}
import dataclasses
from typing import ClassVar
import jax
from jax import lax
import jax.numpy as jnp
import jaxtyping
import numpy as np
import functools
from typing import Protocol, Optional, Self
\end{python}

We define some data types.

\begin{python}
Array = jaxtyping.Array
Float = jaxtyping.Float
PRNGKey = jaxtyping.PRNGKeyArray
Sharding = jax.sharding.Sharding
\end{python}

We recall the implementation of the orthonormal DCT matrices.
\begin{python}[caption=DCT Implemetation, label=snippet:dct]
def dct_type_2_matrix(num_n, num_k, norm="ortho"):
  """Computes the DCT type 2 matrix."""
  n = np.arange(0, num_n)[None, :]
  k = np.arange(0, num_k)[:, None]

  dct_matrix = np.cos(np.pi / num_n * (n + 0.5) * k)

  if norm == "ortho":
    orthogonal_reweigh = np.concatenate([
        np.ones(1) / np.sqrt(num_n),
        np.ones(num_k - 1) * np.sqrt(2 / num_n),
    ])
    dct_matrix = dct_matrix * orthogonal_reweigh[:, None]

  return dct_matrix
\end{python}

Here is an implementation of the \convDCT.
\begin{python}[caption=convDCT Implemetation, label=snippet:conv_dct]
@dataclasses.dataclass(kw_only=True)
class ConvDCT:
  """ConvDCT Transform."""

  block_size: int

  InputT: ClassVar[Float] = Float[Array, "*B H W C"]
  OutputT: ClassVar[Float] = Float[Array, "*B h w F C"]

  def forward(self, x: InputT) -> OutputT:
    """Map to frequency domain."""

    batch_size = x.shape[:-3]

    dct_matrix = dct_type_2_matrix(
        num_n=self.block_size, num_k=self.block_size, norm="ortho"
    )
    dct_kernel = dct_matrix[:, None, :, None] * dct_matrix[None, :, None, :]
    dct_kernel = jnp.reshape(dct_kernel, (-1, 1) + dct_kernel.shape[-2:])

    rx = jnp.reshape(x, (-1,) + x.shape[-3:])

    def convolve(x):
      x = x[..., None]
      return jax.lax.conv_general_dilated(
          lhs=x,
          rhs=dct_kernel,
          window_strides=(1, 1),
          padding="VALID",
          lhs_dilation=(1, 1),
          rhs_dilation=(1, 1),
          dimension_numbers=lax.conv_dimension_numbers(
              x.shape, dct_kernel.shape, ("NHWC", "OIHW", "NHWC")
          ),
      )

    convolved = jax.vmap(convolve, in_axes=-1, out_axes=-1)(rx)
    convolved = jnp.reshape(convolved, batch_size + convolved.shape[1:])
    return convolved

  def inverse(self, x: OutputT) -> InputT:
    """Inverse map."""

    fourier_dim = x.shape[-2]
    if fourier_dim != self.block_size * self.block_size:
      raise ValueError(f"{fourier_dim=} should equal {self.block_size**2}")

    batch_shape = x.shape[:-4]
    height = x.shape[-4] + self.block_size - 1
    width = x.shape[-3] + self.block_size - 1
    out_shape = batch_shape + (height, width, x.shape[-1],)
    # We use the fact that VJP can be used to invert convolutions.
    fwd_primals, vjp_fwd = jax.vjp(
        self.forward, jnp.ones(out_shape, dtype=x.dtype)
    )
    inverted = vjp_fwd(x)[0]
    normalization = vjp_fwd(fwd_primals)[0]

    return inverted / normalization
\end{python}

The following snippet illustrates usage of the \convDCT with some unit tests.
\begin{python}
# Example usage that tests unitarity and inversion
image_shape = (128, 128, 3)
rng = jax.random.PRNGKey(0)
image = jax.random.normal(rng, image_shape)
conv_dct = ConvDCT(block_size=8)
image_f = conv_dct.forward(image)

np.testing.assert_allclose(
        jnp.square(image_f).mean(), jnp.square(image).mean(), rtol=5e-3
    )
convolved = conv_dct.forward(image)
inverted = conv_dct.inverse(convolved)

np.testing.assert_allclose(image, inverted, atol=1e-2, rtol=1e-2)
\end{python}

To implement samplers, we rely on a data structure that represents the noise information and
a Protocol that represents a function that gives the $x$-prediction.

\begin{python}
@jax.tree_util.register_dataclass
@dataclasses.dataclass(frozen=True)
class NoiseInfo:
  """Represents noise information."""
  t: Optional[Array]
  alpha: Array
  sigma: Array
  logsnr: Array

  def broadcast_to(
      self, x: Array, *, sharding: Optional[Sharding] = None
  ) -> Self:
    """Broadcast the noise info to match x's shape and sharding."""
    shape = x.shape
    if sharding is None:
      sharding = jax.typeof(x).sharding

    def fn(y: Array):
      if len(shape) < y.ndim:
        raise ValueError(f'{len(shape)=} shorter than {y.ndim=}')
      return jnp.broadcast_to(
          y.reshape(y.shape + (1,) * (len(shape) - y.ndim)),
          shape,
          out_sharding=sharding,
      )

    return jax.tree.map(fn, self)

class PredictXFn(Protocol):
  """A function that predicts x_t from z_t."""

  def __call__(self, *, z: Array, noise_info: NoiseInfo) -> Array:
    pass
\end{python}

Here is the implementation of our sampler. Note that for the first sampling step ($t=1.0$)
we use a fixed variance because the model is applied out of distribution. This is a small
implementation detail that we found crucial with the SiD2 model but not with the UViT.

\begin{python}[caption=Sampler Implemetation, label=snippet:sampler]
@dataclasses.dataclass(kw_only=True)
class CovAwareSampler:
  """Sampler Implementation.

  first_step_var: variance to use for the first sampling step.
  avg_axes: axes to use to average the estimates in Fourier space.
  block_size: block size for the ConvDCT.
  """

  first_step_var: float = 0.1
  avg_axes: tuple[int, ...] = (-1,)
  block_size: int = 8
  var_cap: float = 1e4

  def fourier(self, dx_dnoise: Array, eps: Array) -> Array:
    """A generalized Hutchinson estimator in Fourier space."""
    dct = ConvDCT(block_size=self.block_size)
    out_sharding = jax.typeof(dx_dnoise).sharding
    # we use auto_axes to avoid adding sharding rules in ConvDCT as Jax figures
    # out sharding correctly
    dct_forward = jax.sharding.auto_axes(dct.forward, out_sharding=out_sharding)
    dct_inverse = jax.sharding.auto_axes(dct.inverse, out_sharding=out_sharding)
    dx_dnoise_f = dct_forward(dx_dnoise)
    eps_f = dct_forward(eps)
    var_f = jnp.mean(eps_f * dx_dnoise_f, axis=self.avg_axes, keepdims=True)
    eps2_f = jnp.mean(jnp.square(eps_f), axis=self.avg_axes, keepdims=True)
    var_f = jnp.clip(var_f, min=0.0, max=self.var_cap) # Numerical stability
    eps2_f = jnp.maximum(eps2_f, 1e-6)  # prevents division by 0.
    x_var_f = jnp.sqrt(var_f / eps2_f)
    eps_f = eps_f * x_var_f
    return dct_inverse(eps_f)

  def predict_noisy_x(
      self,
      *,
      z_t: Array,
      noise_t: NoiseInfo,
      rng: PRNGKey,
      predict_x_fn: PredictXFn,
  ) -> Array:
    # Predict a noised version of x_0 | z_t using covariance estimation
    eps_trace_est: Array = jax.random.normal(rng, z_t.shape, out_sharding=jax.typeof(z_t).sharding)
    l = noise_t.logsnr
    noise_fac = jnp.exp( 
        jnp.minimum(-0.5 * (l + jax.nn.softplus(l)), jnp.log(1e5)) (*@\label{line:noise_fac}@*)
    )
    noise_trace_est = eps_trace_est * noise_fac
    x_pred, dx_dnoise = jax.jvp(
        lambda z: predict_x_fn(z=z, noise_info=noise_t),
        (z_t,),
        (noise_trace_est,),
    )

    dx_dnoise = self.fourier(dx_dnoise=dx_dnoise, eps=eps_trace_est)
    return x_pred + dx_dnoise

  def first_step_noisy_x(
      self,
      *,
      z_t: Array,
      noise_t: NoiseInfo,
      rng: PRNGKey,
      predict_x_fn: PredictXFn,
  ) -> Array:
    eps: Array = jax.random.normal(rng, z_t.shape, out_sharding=jax.typeof(z_t).sharding)
    x_pred = predict_x_fn(z=z_t, noise_info=noise_t)
    return x_pred + eps * jnp.sqrt(self.first_step_var)

  def ddim_step(
    self, *, x: Array, z_t: Array,
    noise_t: NoiseInfo, noise_s: NoiseInfo):


    alpha_s = noise_s.alpha
    alpha_t = noise_t.alpha
    sigma_s = noise_s.sigma
    sigma_t = noise_t.sigma

    sigma_s_div_sigma_t = jnp.where(sigma_s == 0., 0., sigma_s / sigma_t)
    z_s = alpha_s * x + sigma_s_div_sigma_t * (z_t - alpha_t * x)
    return z_s

  def step(
      self,
      *,
      rng: PRNGKey,
      predict_x_fn: PredictXFn,
      z_t: Array,
      noise_s: NoiseInfo,
      noise_t: NoiseInfo,
  ) -> Array:

    first_step = functools.partial(
        self.first_step_noisy_x,
        z_t=z_t,
        noise_t=noise_t,
        rng=rng,
        predict_x_fn=predict_x_fn,
    )
    other_step = functools.partial(
        self.predict_noisy_x,
        z_t=z_t,
        noise_t=noise_t,
        rng=rng,
        predict_x_fn=predict_x_fn,
    )
    noised_x_pred = jax.lax.cond(noise_t.t == 1.0, first_step, other_step)
    noise_t = noise_t.broadcast_to(noised_x_pred)
    noise_s = noise_s.broadcast_to(noised_x_pred)
    return self.ddim_step(
        x=noised_x_pred, z_t=z_t, noise_s=noise_s, noise_t=noise_t
    )
\end{python}

Note that in Snippet~\ref{snippet:sampler} the JVP is applied to the whole prediction function. In the presence of guidance, this would
result in 4 functions evaluations. To prevent the 4th function evaluation, one can use a stop gradient on the unguided $x$-prediction when implementing
\pythoninline{predict_x_fn}.

\section{Samples}
We visualize samples from our ImageNet-512 experiments in Figures~\ref{fig:samples-crane} through \ref{fig:samples-monarchbutterfly}. For each class, we present eight random samples per sampler; to ensure a fair comparison, the seed for the $i$-th sample is kept consistent across all samplers. While the images in this PDF are compressed to maintain a manageable file size, the original high-resolution samples are available in the LaTeX source of this document.

\pgfkeys{
 /samplertable/.is family, /samplertable/.unknown/.style = {/samplertable/default},
 /samplertable/default/.initial = Unknown,
 /samplertable/addim/.initial = aDDIM,
 /samplertable/ddim/.initial = DDIM,
 /samplertable/convdct/.initial = Ours,
 /samplertable/heun/.initial = Heun,
 /samplertable/dpm_solver/.initial = DPM-Solver++
}
\newcommand{\samplerlookup}[1]{\pgfkeysvalueof{/samplertable/#1}}
\pgfkeys{
 /imgcategory/.is family, /imgcategory/.unknown/.style = {/imgcategory/default},
 /imgcategory/default/.initial = Unknown,
 /imgcategory/crane/.initial = Crane (bird).,
 /imgcategory/africanchameleon/.initial = African chameleon.,
 /imgcategory/dragonfly/.initial = Dragonfly.,
 /imgcategory/monarchbutterfly/.initial = Monarch butterfly.

}
\newcommand{\imgcategorylookup}[1]{\pgfkeysvalueof{/imgcategory/#1}}
\newcommand{\tablebody}{}%
\newcommand{\samplestable}[2]{
{
\def\animalname{#1}%
\def\nfe{#2}%
\setlength{\tabcolsep}{1pt}%
\setlength{\extrarowheight}{0pt}%

\def\imagecommand##1##2{
  \includegraphics[width=0.085\textwidth]{images_compressed/\animalname_\nfe_##2_##1.jpg}
  }
\renewcommand{\tablebody}{}%
\foreach \s in {addim, ddim, heun, dpm_solver, convdct} {%
\xdef\tablebody{\tablebody \raise 0.03\textwidth \hbox{\samplerlookup{\s}} &}%
\foreach \n in {1, ..., 7} {%
    \xdef\tablebody{\tablebody \noexpand\imagecommand{\n}{\s}  & }%
}%
\xdef\tablebody{\tablebody \noexpand\imagecommand{8}{\s} \noexpand\\[-3pt]}%
}
\begin{figure}
\begin{tabular}{c *{8}{c|}}
\centering
\renewcommand{\arraystretch}{0}      
\tablebody
\end{tabular}
\caption{Samples on Imagenet512 with \nfe\ NFEs for category: \imgcategorylookup{\animalname}}
\label{fig:samples-\animalname}
\end{figure}
}
}

\samplestable{crane}{36}
\samplestable{africanchameleon}{54}
\samplestable{dragonfly}{84}
\samplestable{monarchbutterfly}{120}

\clearpage

\end{document}